\documentclass[conference]{IEEEtran}
\IEEEoverridecommandlockouts
\usepackage{cite}
\usepackage{amsmath,amssymb,amsfonts}
\usepackage{algorithmic}
\usepackage{tabularx}
\usepackage{graphicx}
\usepackage{textcomp}
\usepackage{xcolor}
\def\BibTeX{{\rm B\kern-.05em{\sc i\kern-.025em b}\kern-.08em
    T\kern-.1667em\lower.7ex\hbox{E}\kern-.125emX}}
\begin{document}

\title{FisHook - An Optimized Approach to Marine Species Classification using MobileNetV2\\}

\makeatletter
\newcommand{\linebreakand}{%
  \end{@IEEEauthorhalign}
  \hfill\mbox{}\par
  \mbox{}\hfill\begin{@IEEEauthorhalign}
}
\makeatother

\author{

\IEEEauthorblockN{Kohav Dey}
\IEEEauthorblockA{\textit{Department of Computer Science and Engineering} \\
\textit{PES University, Banashankari}\\
Bangalore, India \\
deykohav@gmail.com}

\and

\IEEEauthorblockN{Krishna Bajaj}
\IEEEauthorblockA{\textit{Department of Computer Science and Engineering} \\
\textit{PES University, Banashankari}\\
Bangalore, India \\
bajaj.krishna01062002@gmail.com}

\linebreakand

\IEEEauthorblockN{K S Ramalakshmi}
\IEEEauthorblockA{\textit{Department of Computer Science and Engineering} \\
\textit{PES University, Banashankari}\\
Bangalore, India \\
rlsri2305@gmail.com}

\and

\IEEEauthorblockN{Samuel Thomas}
\IEEEauthorblockA{\textit{Department of Electronics and Communication Engineering} \\
\textit{PES University, Banashankari}\\
Bangalore, India \\
samuelthomas1049@gmail.com}

\linebreakand

\IEEEauthorblockN{Sriram Radhakrishna}
\IEEEauthorblockA{\textit{Department of Computer Science and Engineering} \\
\textit{PES University, Banashankari}\\
Bangalore, India \\
sriram.radhakrishna42@gmail.com}

}

\maketitle

\begin{abstract}
Marine ecosystems are vital for the planet's health, but human activities such as climate change, pollution, and overfishing pose a constant threat to marine species. Accurate classification and monitoring of these species can aid in understanding their distribution, population dynamics, and the impact of human activities on them. However, classifying marine species can be challenging due to their vast diversity and the complex underwater environment. With advancements in computer performance and GPU-based computing, deep-learning algorithms can now efficiently classify marine species, making it easier to monitor and manage marine ecosystems. In this paper, we propose an optimization to the MobileNetV2 model to achieve a 99.83\% average validation accuracy by highlighting specific guidelines for creating a dataset and augmenting marine species images. This transfer learning algorithm can be deployed successfully on a mobile application for on-site classification at fisheries.
\end{abstract}


\section{Introduction}
Fish and other seafood have grown to comprise a large part of the world's diet and around 50 percent of the population have become regular consumers of the same. Furthermore, according to the Food and Agricultural Organization of the United Nations, the global consumption of seafood has risen above 20 kilograms per person per year \cite{fao2016global}. In case of a multivariate catch of marine yield, there is a requirement for a highly efficient classification algorithm for the identification of new subspecies with edibility constraints and nutritional composition, as well as a database to add the said species for other pisciculture enthusiasts to utilise. The effective classification of these species can aid in the identification of their rarer counterparts and when combined with market research, it can facilitate informed decision making processes to ensure profitable sales and monitoring of marine species. Regular monitoring helps us understand the distribution, population dynamics, and the impact of human activities on marine species, enabling the development of effective conservation strategies, with conventional methods of classification currently in use being highly effort-intensive, error-prone and time-consuming.

\begin{figure}[htbp]
\centerline{\includegraphics[scale=0.125]{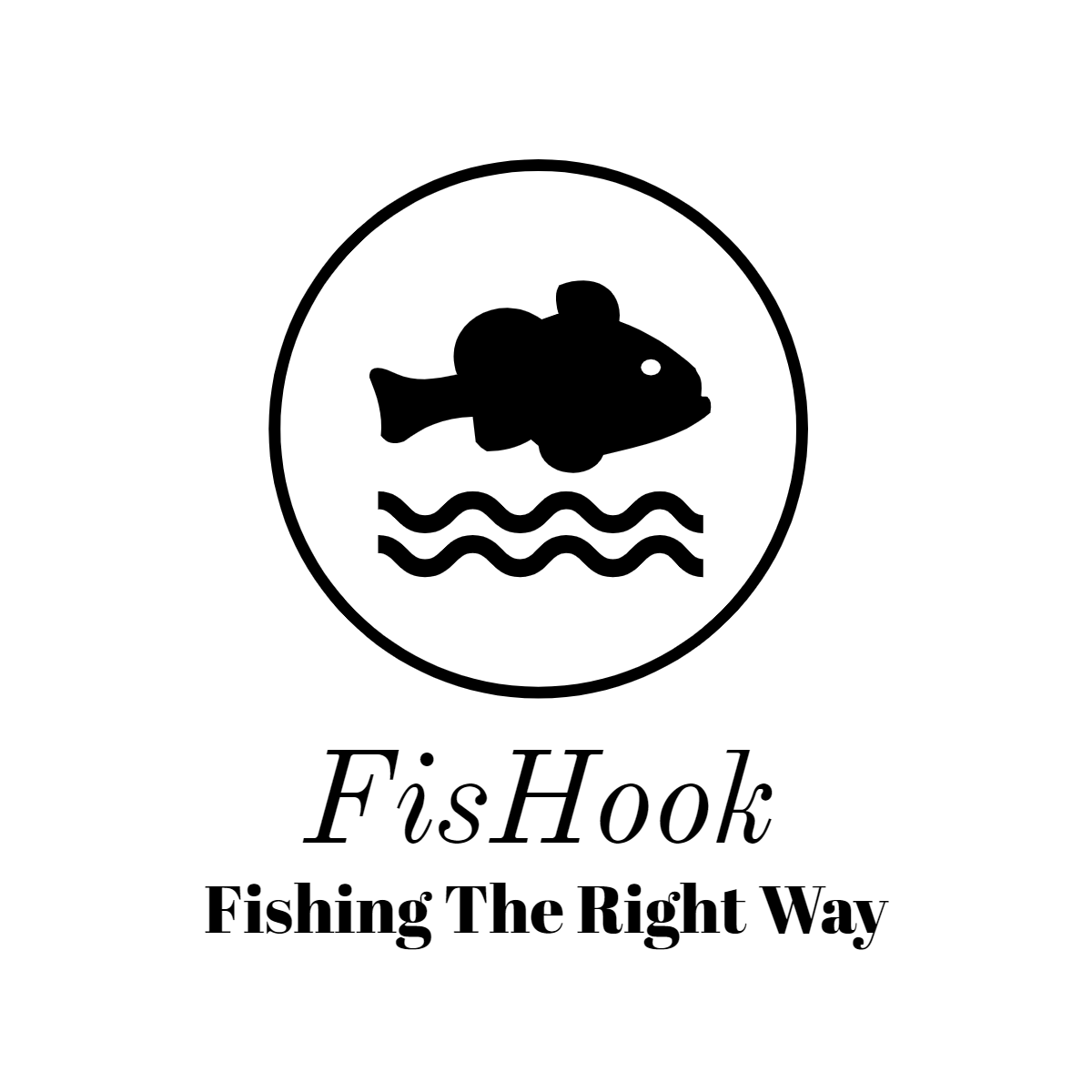}}
\caption{The model introduced in this paper was deployed on an Android mobile application with a mature back-end.}
\end{figure}

\begin{figure}[htbp]
\centering
\centerline{\includegraphics[width=\linewidth]{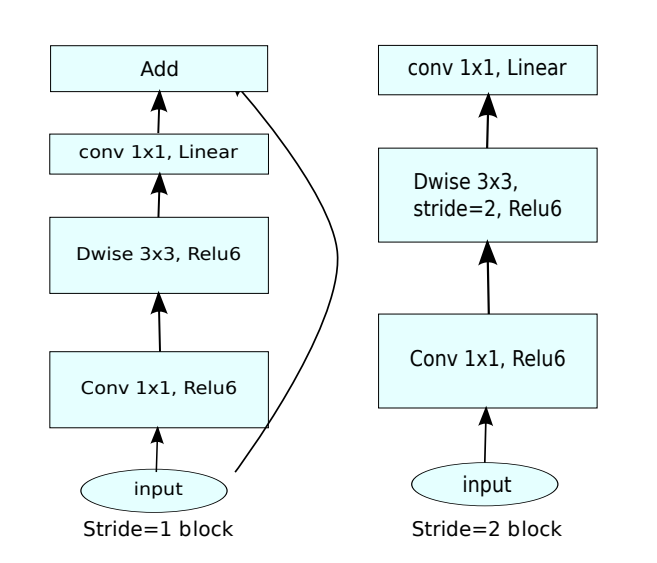}}
\caption{Architecture of MobileNetV2.}
\end{figure}

The proposal aims to develop a marine species classifier using an image-based approach, which will be implemented through transfer learning. To achieve this, we will be utilizing the MobileNetV2 \cite{Sandler_2018_CVPR} architecture, which has been proven to be effective in a variety of image classification tasks. Additionally, we will incorporate a traditional sequential neural network to further enhance the accuracy of the model. The dataset used for training and testing the model consists of nine distinct fish classes \cite{ulucan2020large}, each with its unique characteristics and features. By leveraging transfer learning, we are utilising the pre-trained MobileNetV2 model initialised with ImageNet \cite{kornblith2019better} weights, as a feature extractor. This will enable the model to leverage the pre-existing knowledge of the MobileNetV2 architecture to improve the classification accuracy of the dataset. The primary objective of this marine species classifier is to accurately identify and differentiate between the various fish classes in the dataset. This classifier can be utilized in a range of marine applications, such as monitoring and conservation efforts, to better understand the distribution and behavior of different fish species.

\section{Related Work}
At present, several machine learning models trained and tested on various datasets have been implemented to achieve an acceptable classification accuracy. Ulucan et al. \cite{ulucan2020large} implemented a feature-based classification model using SVMs providing an average training and validation accuracy of 98.01 percent and 88.69 percent, for 9 distinct classes. Rossi et al. \cite{rossi2016fishapp} built a mobile application with a remote server for cloud-based machine learning, which implements an Artificial Neural Network(ANN) classifier, with an accuracy of 100 percent, for 7 distinct fish classes. Cchabra et al. \cite{chhabra2020hybrid} built a VGG16 transfer learning model achieving an accuracy of 93.8 percent for 8 different classes of fish. Liu et al. \cite{liu2019real} proposed a MobileNetV2 based transfer learning model, achieving an accuracy of 95 percent, for 7 categories of fishes. The optimized approach proposes an implementation that contributes to the latter, by specifying guidelines for the creation and augmentation of a dataset as well as changes to the existing model to obtain an improvement in the accuracy of the model. \\

This literature survey indicates that MobileNetV2-based deep learning models have been successfully applied for fish species identification and classification tasks with high accuracy. The studies reviewed in this survey have used different techniques such as transfer learning, data augmentation, feature extraction, and hybrid models to improve the performance of the MobileNetV2-based models. The proposed research paper can build upon the existing literature by proposing a novel approach for fish species identification using MobileNetV2 and exploring the potential of different techniques to improve the accuracy of the model.

\section{Concept Theory and Implementation}

Marine life is essential to the planet's ecosystem, providing food, oxygen and regulating climate change. As it stands, innumerable species currently live under constant threat from climate change, pollution, overfishing, and other human activities. Accurate classification and monitoring of marine species can help us understand their distribution, population dynamics, and the impact of human activities on them, allowing us to develop effective conservation strategies. The classification of marine species can be a challenging task due to the vast diversity of species and the complex underwater environment. With improvements in computer performance and advancements in GPU-based computing, it has become possible to classify marine species accurately and efficiently using deep-learning algorithms to process large quantities of data, making it easier to monitor and manage marine ecosystems. In this paper, we present an optimization to the most accurate image classification model for the purposes of marine specie classification. The issue has been evaluated in the past using the MobileNetV2 model and has already yielded a significant average classification accuracy of 97.26 percent \cite{fitrianah2022fine}. MobileNetV2 is a neural network architecture that achieves state-of-the-art performance on a range of tasks, including image classification, object detection, and semantic segmentation. Moreover, MobileNetV2 is computationally efficient, achieving a speedup of up to 3.4x over its predecessor, MobileNetV1, while maintaining similar accuracy. However, we noticed that the model can be significantly improved upon by introducing modifications to the type of dataset used for training. By highlighting specific guidelines for the creation of such a dataset  we've achieved an improved validation accuracy average of 99.83 percent, while at the same time, appending it with augmented versions of the same marine specie images along with the enhanced model was studied and found to hinder the effectiveness of the model. The transfer learning algorithm was implemented with the use case of on-site classification at fisheries and deployed successfully on a mobile application.

\subsection{Datasets}
A dataset consisting of 9 species of marine fauna, was used for the training a modified version of the classifier implemented by Ulucan et. al \cite{ulucan2020large} Each of the seven classes - red mullet, gilt head bream, horse mackerel, sea bass, red sea bream, black sea sprat and striped red mullet, consisted of 50 distinct fish images, while the other two - trout and shrimp, had 30 distinct images. The dataset and the images captured were varied enough to capture all possible physical details of the subjects by virtue of them being taken from multiple angles to the surface as well as to the body-axis of the subjects themselves. The images were sized at 590 x 445 pixels. The images were passed on to an augmentation algorithm where each image is rotated with non-repeated random angles and they are reflected, while maintaining a consistent image size. Dataset collection and augmentation concluded with each seafood type consisting of approximately 1,100 images.

\begin{figure}[htbp]
\centering
\centerline{\includegraphics[width=\linewidth]{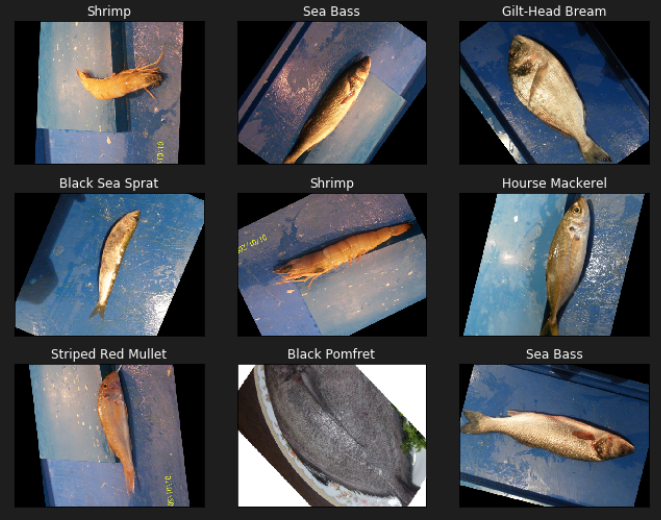}}
\caption{Sample images along with labels in the training dataset after augmentation.}
\end{figure}

Several approaches were undertaken to build the classifier. Any TensorFlow based model which has to be run as an application in edge devices, needs to be converted to a compressed .tflite model. Of all, MobileNetV2 delivered promising prediction results, following which the Tensorflow \cite{abadi2016tensorflow} model was converted into a .tflite \cite{shah2020multi} model format to be deployed in a Flutter application. The paper gives an end-to-end implementation of the mentioned approach, a brief overview of the various shortcomings identified and the future scope for improvement as well.

\begin{figure}[htbp]
\centering
\centerline{\includegraphics[width=\linewidth]{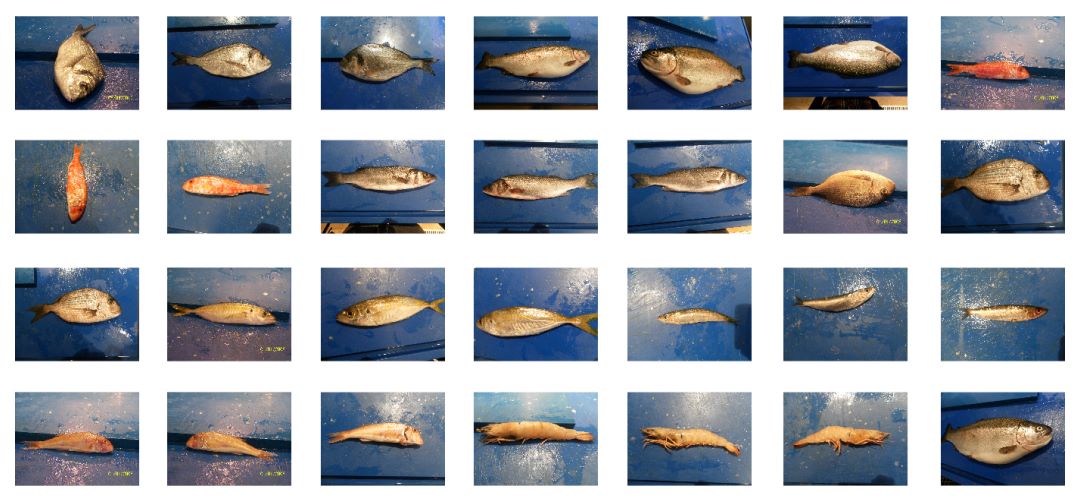}}
\caption{The original dataset that the model was trained on \cite{ulucan2020large}.}
\end{figure}

\subsection{Image Preprocessing}
Augmentation of the dataset was performed by the ImageDataGenerator \cite{arora2021analytical} which generates batches of tensor image data with real-time data augmentation \cite{devries2017dataset}. Augmentation and pre-processing was done implicitly with a validation split of 0.2 in the training dataset. The pre-process-input API scales the input pixels between -1 and 1. Tensors of varied dimensions are reshaped to 224X224 pixels, as an appropriate input to the MobileNetV2 network for feature extraction followed by, classification by CNN.

\subsection{Transfer Learning}
Transfer learning \cite{weiss2016survey} is a machine learning technique which focuses on reusing pre-trained models for similar deep learning tasks, given the vast compute resources and the time taken to train such complex neural networks on massive amounts of data. Since most real-world problems do not have sufficient labelled data points that can be used for training purposes, this kind of an optimisation enables the training of deep neural networks with comparatively less data, alongside reducing the training time and improving the performance of the model. 

A base network is trained on a base dataset and task, after which the learned features are repurposed or transferred to a second target network, which is trained for a different task, on a different target dataset with similar high-level features to the original. Transfer learning produces better results when the features learnt by the model are general to both the base and target dataset. This type of transfer learning is called as inductive transfer.

\subsection{MobileNetV2 - Overview}
MobileNetV2 \cite{Sandler_2018_CVPR}, introduced by Google is a convolutional neural network architecture that is designed to perform efficiently on edge devices. It is based on an inverted residual structure where the residual connections are between the bottleneck layers, which is an improvisation over MobileNetV1. MobileNetV1 uses depth-wise separable convolution and has a reduced complexity cost as well. The intermediate expansion layer uses lightweight depth-wise convolutions to filter features as a source of non-linearity in the narrow layers. 

On a whole, the architecture of MobileNetV2 containing the initial fully convolution layer with 32 filters, followed by 19 residual bottleneck layers is used for feature extraction in the implementation to generate a feature map as an input to the fully connected Dense layers of the traditional CNN. Bottleneck layers are added to reduce the number of feature maps in the network, which, otherwise, tend to increase in each layer. This is achieved by using 1x1 convolutions with fewer output channels than input channels, thereby reducing the model complexity and make it light-weight that can be implemented in edge devices, like smartphones.

\subsection{Model Architecture}

The paper proposes a transfer learning based implementation using MobileNetV2 where the fully convolutional pre-trained layers of the model are used for feature extraction and a traditional neural network with fully connected Dense layers are used for classification. MobileNetV2 takes an input tensor of size 224 x 224 x 3 in the input layer. The weights of the top-trainable layers are frozen and the generated feature map is fed as an input to the fully connected dense layers of the neural network. Non-linearity is introduced in the hidden layers of the network by the RELU activation function, which allows the stacking of multiple layers of neurons to create a deep neural network. Non-linearity helps the network to learn complex datasets, which high accuracy.

\begin{equation} 
Relu(x) = max(0, x)\\
\end{equation}

\begin{figure}[htbp]
\centering
\centerline{\includegraphics[width=\linewidth]{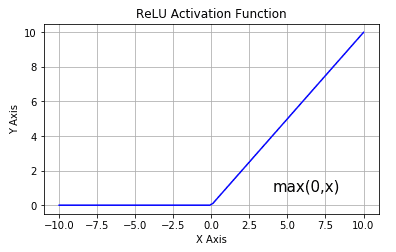}}
\caption{RELU Activation Function for reference.}
 
\end{figure}

Figure 5 is a graphical representation of the RELU function (Equation 1) which produces the same output as the input for non-zero values and all negative values are reduced to zero.

Every layer in the neural network is trained with an associated loss function, to quantify the performance of the model during the training phase. It is a measure of the difference between the predicted output and the actual output for a given input, which is used to optimize the model's parameters in the subsequent training iterations. Minimizing the loss function can significantly improve the accuracy of the model and make it more effective at its intended task.

Categorical cross-entropy loss \cite{zhang2018generalized} is a commonly used loss function in deep learning for multi-class classification problems.

\begin{equation}   
-{(y\log(p) + (1 - y)\log(1 - p))}
\end{equation}

Loss for a single class label per observation is given by Equation 3.

\begin{equation}   
-\sum_{c=1}^My_{o,c}\log(p_{o,c})
\end{equation}
If M is greater than 2 (i.e. multiclass classification), loss for each class label per observation is calculated separately and added to the result cumulatively. The resulting sum gives the categorical cross entropy loss for the classifier (Equation 4).

The final output layer uses the softmax activation for better intepretation of classification, scaling the prediction of each class in the numpy array to either 0 or 1. Transfer learning with MobileNetV2 is used to classify different species of fishes and the method produces acceptable results since the base neural network has learnt the features necessary to classify fishes from the ImageNet dataset.


\begin{equation}   
\sigma(z_i) = \frac{e^{z_{i}}}{\sum_{j=1}^K e^{z_{j}}} \ \ \ for\ i=1,2,\dots,K
\end{equation}

Equation 2 is a mathematical representation of the softmax \cite{dunne1997pairing} function, which normalises the prediction probabilities of various classes in a multi-class classifier.

During the training phase, measuring accuracy is an essential part of evaluating the performance of deep learning models on both the training dataset and validation dataset. The following are the commonly used metrics used to evaluate the model's performance during training:


1. Training accuracy: It measures the accuracy of the model on the training dataset during the training process. Training accuracy is measured as a percentage of correctly classified instances in the training dataset.

\begin{equation}   
Accuracy_{training} = \frac{1}{n}\sum_{i=1}^n (y_{i} - \hat{y_{i}}) \\
\end{equation}

2. Training loss: It measures the error or difference between the predicted output and the actual output on the training dataset during the training process. The training loss is used to optimize the model's parameters to improve its performance.

\begin{equation}   
L_{train} = \frac{1}{n}\sum_{i=1}^n L(y_{i}, \hat{y_{i}}) \\
\end{equation}

3. Validation accuracy : It measures the accuracy of the model on the validation dataset during the training process. The validation accuracy provides an estimate of how well the model will perform on new data.

\begin{equation}   
Accuracy_{valdation} = \frac{1}{m}\sum_{i=1}^m (y_{i} - \hat{y_{i}}) \\
\end{equation}

4. Validation loss : It measures the error or difference between the predicted output and the actual output on the validation dataset during the training process. The validation loss is used to monitor overfitting, where the model performs well on the training dataset but not on the validation dataset.

\begin{equation}   
L_{val} = \frac{1}{m}\sum_{i=1}^m L(y_{i}, \hat{y_{i}}) \\
\end{equation}

During the training process, we aim to minimize the training loss and improve the training accuracy while avoiding overfitting. Validation accuracy and validation loss are used to monitor the model's performance and prevent overfitting. If the validation loss increases while the training loss decreases, it indicates that the model is overfitting and needs to be adjusted.

\section{Deployment}

The use of a mobile application for data collection, coupled with an AWS/GCP based deployment for processing and classification, is a promising approach to building a marine species classifier. In this section, we discuss the design and implementation of a Flutter-based mobile application and its integration with the cloud server.

The cross-platform mobile application was developed using Flutter and was designed to be user-friendly, accessible and customizable. The application allowed users to upload/capture pictures of marine species and then sent to a Firebase shared storage, from where the images were requested by an EC2 instance for processing and classification\cite{juve2009scientific}.

The AWS pipeline was designed to efficiently process large volumes of data in real-time. The algorithm was deployed using AWS' EC2 service, an elastic computation unit on the cloud that provided scalable, reliable, and cost-effective model training and deployment with the aid of purpose-built automation scripts. The model was updated continually with new data as and when the user feeds for classification, ensuring that the algorithm remained accurate and relevant over time.

Pipelining the process in this manner allows the distribution of the computational load exerted by the training process across multiple remote computers \cite{zomaya1996parallel} at the point of exchange from the shared Firebase storage \cite{moroney2017firebase} (being used as a bucket in this implementation). This would greatly increase the cap on throughput and assist any extended user-base for the mobile application by ensuring scalability.

\begin{figure}[htbp]
\centerline{\includegraphics[scale=1]{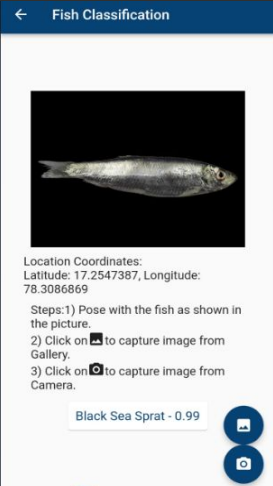}}
\caption{An example of an image taken from the dataset itself, being evaluated by the model accurately.}
 
\end{figure}

The mobile application provided a simple and accessible way for users to collect data, while the cloud pipeline ensured the efficient processing and classification of this data. The integration of a Flutter-based mobile application with AWS represented a powerful and innovative approach to marine species classification. This approach has the potential to contribute to more effective marine conservation efforts by enabling more accurate and efficient data collection and classification.

\section{Results}


\subsection{Testing Specifications}
TensorFlow, a popular open-source library developed by Google for machine learning and artificial intelligence, was used to conduct the experiment along with the use of an additional supporting package known as Keras \cite{gulli2017deep} for further code abstraction. The experimental setup used to train the model prior to deployment adopts an Ubuntu 20.04 version computer, with specifications including the 8-core, 16-thread, AMD Ryzen 9 5900S processor, 16GB DDR4 RAM, 1TB disk memory and a dedicated NVIDIA GeForce RTX 3060 GPU, with 6GB VRAM.

\subsection{Training Process}
The model uses Categorical Cross Entropy loss function to find the error or deviation in the learning process and the Adam optimizer to optimize the input weights by comparing the prediction and loss function, used in the training process.

Training accuracy, training loss, validation accuracy and validation loss are the metrics used to evaluate the performance of the model. The neural network was trained with a batch size of 32, for 50 epochs with early stopping callback, where the training process terminates when there is an increase in the validation loss of the model. Feature extraction is done by the pre-trained layers of the MobileNetV2 model and further an artificial neural network consisting of two fully connected dense layers has been appended to perform multi-class fish classification. The model was trained for 312 seconds 406 milliseconds (5.223 minutes) over 3 epochs, with approximately 104 seconds 468 milliseconds per epoch.


\subsection{Training \& Validation Results}
For reference, the training accuracy, training loss, testing accuracy and testing loss are quantitatively analyzed in Table 1.
\\

\begin{table}[!h]
\begin{center}
\resizebox{8cm}{5cm}{
    \begin{tabular}{|p{1.8cm}|p{1.2cm}|p{1.2cm}|p{1.2cm}|p{1.2cm}|}
        \hline
        \multicolumn{5}{|c|}{Results}\\
        \hline
        Model & Training Accuracy & Training Loss & Validation Accuracy & Validation Loss\\
        \hline
        InceptionV3 and custom non-augmented dataset based underwater & 99.73\% & 0.0072 & 91.13\% & 0.372\\
        MobileNetV1 and custom non-augmented dataset based underwater & 99.29\% & 0.0199 & 89.92\% & 0.5113\\
        MobileNetV2 and custom non-augmented dataset based underwater & 99.58\% & 0.0085 & 92.89\% & 0.3165\\
        MobileNetV2 with ANN and custom non-augmented dataset based outside water & 99.99\% & 0.0021 & 99.83\% & 0.0051\\
        \hline
    \end{tabular}
} 
\break
\break
\caption{A quantitative analysis of the test results.}
\end{center}
\end{table}

\begin{figure}[htbp]
\centerline{\includegraphics[trim={0 0.4cm 0 0}, clip]{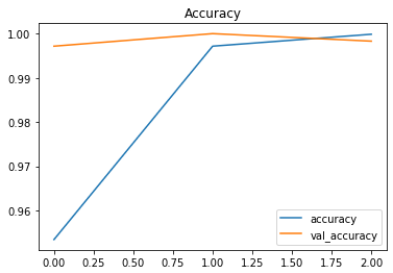}}
\caption{Accuracy plots of the training and validation phase.}
 
\end{figure}

\begin{figure}[htbp]
\centerline{\includegraphics[trim={0 0.35cm 0 0}, clip]{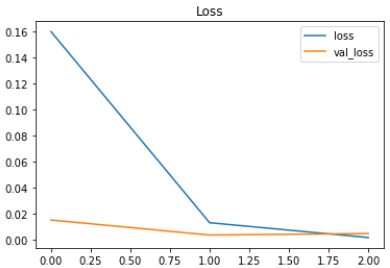}}
\caption{Loss plots of the training and validation phase.}
 
\end{figure}

From the graph curves in Figure 7 and 8, it can be observed that the validation accuracy and loss increase with an increase in number of iterations in the network training process, and it stabilises in a certain interval. With the nature of the trajectory, it can be inferred that the model was trained efficiently with appropriate hyper-parameters to produce an acceptable accuracy and loss, with very minimal over-fitting.

Drawing parallels to the results obtained by Liu et al. \cite{liu2019real} , the mean accuracy of 99.83\% achieved with the CNN model based on MobileNetV2 architecture is an improvisation to the model proposed with an SVM based classifier, trained on the same dataset, which provides a mean accuracy of 98.01\%. 
The trained model was then converted into a special efficient portable Flatbuffer, the Tensorflow Lite format (.tflite) to be deployed in edge devices. The model was imported in a cross platform Flutter framework application, and  was used to predict an unseen image (Figure 6), which classified the fish as a Black Sea Sprat in this example, with a confidence measure of 0.99. 

Interestingly enough, it must be noted here that when the same model was trained on the previously mentioned augmented dataset, it brought the validation accuracy of the model far below 50\%. This was because the original dataset by Ulucan et al. \cite{ulucan2020large} was already implicitly augmented as it contained images of the same fish captured at multiple angles, allowing the dataset itself to successfully capture unique low-level features \cite{khan2021image} of the fish in its entirety. Therefore, further augmentation introduced a large amount of noise into the dataset, causing the network to constantly change its weights throughout the training phase without significant convergence.


\subsection{Deployment Accuracy}

\begin{figure}[htbp]
\centerline{\includegraphics[scale=0.27]{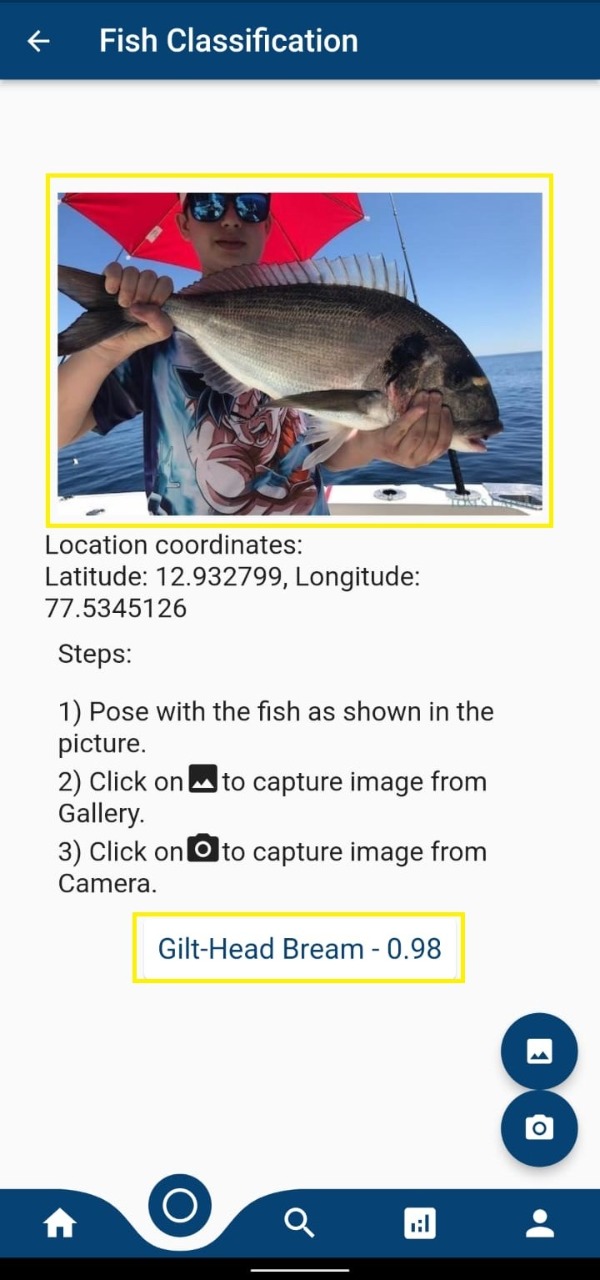}}
\caption{An example of an image with significant noise evaluated successfully by the model\cite{komen}.}
\end{figure}

While the training and testing of the model yielding promising results on the dataset provided by Ulucan et al. \cite{ulucan2020large}, it's important to note here that the model is able to make accurate predictions even upon encountering images that never featured in said dataset, as well as those that additionally contained noise. Refer to figure 9 for one such example.

It can be observed in this demo that the picture evaluated by the model contained plenty of noise in the form of a human holding the Gilt-Head Bream and occlusion of the tail joint as well as the region below the gills. \\
It would be fair to claim at this point that the model produced a good one-off prediction. However, upon testing it with 38 more such images sourced publicly, it was observed that the model accurately classified 35 images with a confidence of 0.5 or more, yielding a 92.10\% deployment accuracy. 

The most reasonable hypothesis that can be drawn at this point is that the strong lighting in the original dataset highlighted the stray marks on the cutting board well enough for them to be picked up as low level features by the neural network, hence training the model to account for significant noise in its predictions as well

\section{Conclusion and End Notes}

In conclusion, the algorithm presented in this research paper has demonstrated impressive accuracy in classifying multiple marine species based on data collected through a mobile application. With a reported testing accuracy of 99.83\%, this algorithm has the increased potential to play a crucial role in marine conservation efforts with the improvements acheived upon the status quo.

The implementation of the algorithm in a server-side framework on AWS allows for continual updates and improvements to the model based on new data, which is particularly important in a field as dynamic as marine biology. The research has the potential to make significant contributions to ongoing efforts to better understand and conserve marine ecosystems and provides a promising step towards more accurate and efficient methods for marine species classification. Further development and refinement of this algorithm could have far-reaching implications for the field of marine conservation, helping to ensure the preservation and protection of marine species for generations to come.


\bibliography{references.bib}
\bibliographystyle{IEEEtran}\

\end{document}